# Structural Group Sparse Representation for Image Compressive Sensing Recovery

Jian Zhang*, Debin Zhao*, Feng Jiang*, and Wen Gao*+

*School of Computer Science & Technology, Harbin Institute of Technology, China
+School of Electronic Engineering & Computer Science, Peking University, China
{jzhangcs, dbzhao, fjiang}@hit.edu.cn, wgao@pku.edu.cn

**Abstract:** Compressive Sensing (CS) theory shows that a signal can be decoded from many fewer measurements than suggested by the Nyquist sampling theory, when the signal is sparse in some domain. Most of conventional CS recovery approaches, however, exploited a set of fixed bases (e.g. DCT, wavelet, contourlet and gradient domain) for the entirety of a signal, which are irrespective of the nonstationarity of natural signals and cannot achieve high enough degree of sparsity, thus resulting in poor rate-distortion performance. In this paper, we propose a new framework for image compressive sensing recovery via structural group sparse representation (SGSR) modeling, which enforces image sparsity and self-similarity simultaneously under a unified framework in an adaptive group domain, thus greatly confining the CS solution space. In addition, an efficient iterative shrinkage/thresholding algorithm based technique is developed to solve the above optimization problem. Experimental results demonstrate that the novel CS recovery strategy achieves significant performance improvements over the current state-of-the-art schemes and exhibits nice convergence.

## 1. Introduction

Compressive Sensing (CS) has drawn quite an amount of attention as an alternative to the current methodology of sampling followed by compression [1–2]. By exploiting the redundancy existed in a signal, CS conducts sampling and compression at the same time. From many fewer acquired measurements than suggested by the Nyquist sampling theory, CS theory demonstrates that, a signal can be reconstructed with high probability when it exhibits sparsity in some domain.

An attractive strength of CS-based compression is that the encoder is made signal-independent and computationally inexpensive at the cost of high decoder complexity, that is, simple encoder and complex decoder. This asymmetric design is severely desirable in some image processing applications when the data acquisition devices must be simple (e.g. inexpensive resource-deprived sensors), or when oversampling can harm the object being captured (e.g. X-ray imaging) [3].

From CS theory, the sparsity degree of a signal plays a significant role in recovery. The higher degree of a signal, the higher recovery quality it will have. Thus, seeking a domain in which the signal has a high degree of sparsity is one of the main challenges which CS recovery should face. Since natural signals such as images are typically nonstationary, there exists no universal domain in which all parts of the signals are sparse. The most current CS recovery methods explored a set of fixed domains (e.g. DCT, wavelet and contourlet, gradient domain) [17, 8, 19], therefore are signal-independent or not adaptive, resulting in poor rate-distortion performance.

To rectify the problem, many recent works incorporated additional prior knowledge about transform coefficients (statistical dependencies, structure, etc.) into the CS recovery framework, such as Gaussian scale mixtures (GSM) models [9], tree-structured wavelet [4], tree-structured DCT (TSDCT) [7]. Additionally, in [6], a projection-driven CS recovery coupled with block-based random image sampling is developed, which aims to encourage sparsity in the domain of directional transforms. Chen *et al.* [18] exploited multi-hypothesis predictions to generate a residual in the domain of the CS random projections, where this residual being typically more compressible than the original signal leads to improved reconstruction quality. Furthermore, many latest works concentrate on utilization of both local and nonlocal statistics for high quality image restoration [5, 16]. Zhang *et al.* [21, 10] proposed a framework for CS recovery via collaborative sparsity, which enforces local 2-D sparsity and nonlocal 3-D sparsity simultaneously in an adaptive hybrid space-transform domain, thus greatly confining the CS solution space.

In recent years, very impressive image restoration results have been obtained with local patch-based sparse representations calculated with dictionaries learned from natural images [13], [15]. Relative to fixed dictionaries, learned dictionaries enjoy the advantage of being better adapted to the images, thereby enhancing the sparsity. However, dictionary learning is a large-scale and requires high computational complexity. What's more, in the process of dictionary learning, each patch is considered independently, which ignores the relationships between similar patch, such as self-similarity [13].

Considering the fact that the natural image signal is non-stationary and its sparse domain varies spatially, in this paper, we establish a new sparse representation modeling, called structural group sparse representation (SGSR), and develop a novel strategy for image compressive sensing recovery via SGSR. The proposed SGSR enforces image sparsity and self-similarity simultaneously in a unified framework, which enables a natural image to be highly sparse in an adaptive group domain. To make SGSR tractable and robust, an iterative shrinkage/thresholding algorithm (ISTA) based technique is developed to solve the above severely underdetermined inverse problem efficiently. Extensive experiments manifest that SGSR is able to increase recovery quality by a large margin compared with the conventional CS recovery methods or require many fewer measurements for a desired reconstruction quality.

The remainder of the paper is organized as follows. Section 2 reviews CS theory and traditional patch based sparse representation modeling. Section 3 provides the design of structural group sparse representation modeling. Section 4 gives the implementation details of proposed CS recovery strategy via SGSR. Experimental results are reported in Section 5. In Section 6, we conclude this paper.

## 2. Background

### 2.1. Compressive Sensing

A signal $u$ of size $N$ is said to be sparse in domain or basis $\Psi$, if its transform coefficients $\alpha = \Psi^T u$ are mostly zeros, or nearly sparse if the dominant portion of coefficients are either zeros or very close to zeros. The sparsity of $u$ in $\Psi$ is quantified by the number of significant elements within the coefficients vector $\alpha$.

More specifically, given $M$ linear measurements, the CS recovery of $u$ from $b$ is formulated as the following constrained optimization problem:

$$\min_{\boldsymbol{u}} \left\| \Psi^T \boldsymbol{u} \right\|_p \quad \text{s.t.} \quad \boldsymbol{b} = \boldsymbol{A}\boldsymbol{u}, \tag{1}$$

where $\boldsymbol{A}$ represents the random projections (RS). $p$ is usually set to 1 or 0, characterizing the sparsity of the vector $\Psi \boldsymbol{u}$. $\|*\|_1$ is $\ell_1$ norm, adding all the absolute values of the entries in a vector, while $\|*\|_0$ is $\ell_0$ norm, counting the nonzero entries of a vector. According to [2], CS is capable of recovering $K$-sparse signal $\boldsymbol{u}$ (with an overwhelming probability) from $\boldsymbol{b}$ of size $M$, provided that the number of random samples meets $M \geq cK(N/K)$. The required sampling rate ($M/K$), to incur lossless recovery, is roughly proportional to ($K/N$). A compressive imaging camera prototype using RS has been presented in [20].

## 2.2. Traditional Patch based Sparse Representation

Recently, sparse representation based modeling has been proven to be a promising model for image processing tasks [13, 15], which assumes that natural image is sparse in some domain spanned by a set of bases or a dictionary of atoms. In literature, the basic unit of sparse representation for images is patch. Mathematically, denote by $\boldsymbol{x} \in \mathbb{R}^N$ the original image, and by $\boldsymbol{x}_i$ an image patch of size $\sqrt{B_s} \times \sqrt{B_s}$ at location $i$, $i = 1, 2, ..., n$. Given a dictionary $\boldsymbol{D} \in R^{B_s \times M}$, each patch can be sparsely coded as $\boldsymbol{\alpha}_i = \boldsymbol{D}^T \boldsymbol{x}_i$ by using some sparse coding algorithms. Then the entire image can be sparsely represented by the set of sparse codes $\{\boldsymbol{\alpha}_i\}$. Therefore, the CS recovery problem under the dictionary $\boldsymbol{D}$ is formulated below:

$$\min_{\boldsymbol{x}} \sum_{k=1}^{n} \left\| \boldsymbol{D}^T \boldsymbol{x}_k \right\|_p \quad \text{s.t.} \quad \boldsymbol{b} = \boldsymbol{A}\boldsymbol{x}. \tag{2}$$

## 3. Structural Group Sparse Representation Modeling

Besides sparsity, nonlocal self-similarity is another significant property of natural images, leading to great success in image restoration [11, 12]. It characterizes the repetitiveness of the textures and structures embodied by natural images within nonlocal area, which can be used for retaining the sharpness and edges effectually to maintain image nonlocal consistency.

To resolve the problem of CS recovery, in this paper, we propose a novel sparse representation modeling in the unit of group instead of patch, to exploit the sparsity and the self-similarity of natural images simultaneous in a unified framework. Each group is composed of nonlocal patches with similar structures as its columns. Thus, the proposed sparse representation modeling is named as structural group sparse representation (SGSR). Moreover, adaptive dictionaries for each group can also be learned to achieve sparser representation with very low complexity, which enables the proposed SGSR more efficient and effective. This section will give detailed description of SGSR modeling, and the adaptive dictionary learning technique will be provided in the next section.

First, divide the image $\boldsymbol{x}$ with size $N$ into $n$ overlapped patches of size $\sqrt{B_s} \times \sqrt{B_s}$ and each patch is denoted by $\boldsymbol{x}_k$, i.e., $k = 1, 2, ..., n$. Then, for each patch $\boldsymbol{x}_k$, in the $L \times L$ training window, search its $c$ best matched patches, which consist of the set $\boldsymbol{S}_{\boldsymbol{x}_k}$. Next, define the structural group $\boldsymbol{G}_{\boldsymbol{x}_k}$ corresponding to $\boldsymbol{x}_k$, which includes every element of $\boldsymbol{S}_{\boldsymbol{x}_k}$ as its columns, i.e., $\boldsymbol{G}_{\boldsymbol{x}_k} = \{\boldsymbol{G}_{\boldsymbol{x}_k \otimes 1}, \boldsymbol{G}_{\boldsymbol{x}_k \otimes 2}, ..., \boldsymbol{G}_{\boldsymbol{x}_k \otimes c}\}$. Finally, given the adaptive dictionary $\boldsymbol{D}_k$ for each group, $\boldsymbol{G}_{\boldsymbol{x}_k}$ can be sparsely represented as $\boldsymbol{\alpha}_k = \boldsymbol{D}_k^T \boldsymbol{G}_{\boldsymbol{x}_k}$. Therefore, the CS recovery

problem with the proposed structural group sparse representation (SGSR) modeling is formulated as:

$$\min_{x} \sum_{k=1}^{n} \left\| D_k^T G_{x_k} \right\|_p \quad \text{s. t.} \quad b = Ax. \tag{3}$$

In this paper, $p$ is set to be 0 to achieve higher sparse representation. Let $\alpha_k = D_k^T G_{x_k}$ and $\alpha_x$ denotes the concatenation of all $\alpha_k$, that is, $\alpha_x = [\alpha_1^T, \alpha_2^T, ..., \alpha_n^T]^T$. Thus, the CS recovery problem (3) is rewritten as

$$\min_{x} \left\| \alpha_x \right\|_0 \quad \text{s. t.} \quad b = Ax. \tag{4}$$

## 4. Numerical Algorithm Implementation Details

In this section, we will present the implementation details for solving Problem (4) for image CS recovery. Note that Problem (4) is essentially non-convex and quite difficult to solve directly due to the non-differentiability and non-linearity of the SGSR term. Solving it efficiently is one of the main contributions of this paper.

We first introduce a penalty parameter $\lambda$ and transform Problem (4) into the following unconstrained form:

$$\min_{x} \tfrac{1}{2} \left\| Ax - b \right\|_2^2 + \lambda \left\| \alpha_x \right\|_0, \tag{5}$$

where $\lambda$ is the regularization parameter controlling the trade-off between two terms in Eq. (5). Then, invoking iterative shrinkage/thresholding algorithms (ISTA) [14] to Eq. (5) leads to the following two iterative steps:

$$r^{(j)} = x^{(j)} - \rho A^T (Ax^{(j)} - b), \tag{6}$$

$$x^{(j+1)} = \underset{x}{\operatorname{argmin}} \tfrac{1}{2} \left\| x - r^{(j)} \right\|_2^2 + \lambda \left\| \alpha_x \right\|_0, \tag{7}$$

where $\rho$ is a constant stepsize and $j$ denotes the iteration number. Hence, the key for solving Eq. (5) is to solve Eq. (7) efficiently. For simplicity, the subscript $j$ is omitted without confusion.

Note that it is difficult to solve Eq. (7) directly due to the complicated definition of $\alpha_x$. To enable solving Eq. (7) tractable, in this paper, a general assumption is made, with which even a closed form can be achieved. Concretely, $r$ can be regarded as some type of the noisy observation of $x$, and then an assumption is made that each element of $x$-$r$ follows an independent zero-mean distribution with variance $\sigma^2$. It is worth emphasizing that the above assumption does not need to be Gaussian process, which is more general and reasonable. By the general assumption, we can prove the following conclusion.

**THEOREM 1.** *Let $x, r \in \mathbb{R}^N, G_{x_k}, G_{r_k} \in \mathbb{R}^{B_s \times c}$, and denote the error vector by $e = x - r$ and each element of $e$ by $e(j)$, $j = 1,...,N$. Assume that $e(j)$ is independent and comes from a distribution with zero mean and variance $\sigma^2$. Then, for any $\varepsilon > 0$, we have the following property to describe the relationship between $\left\| x - r \right\|_2^2$ and $\sum_{k=1}^{n} \left\| G_{x_k} - G_{r_k} \right\|_2^2$, that is,*

$$\lim_{N \to \infty, K \to \infty} P\left\{ \left| \left\| x - r \right\|_2^2 - \tfrac{N}{K} \sum_{k=1}^{n} \left\| G_{x_k} - G_{r_k} \right\|_2^2 \right| < \varepsilon \right\} = 1, \tag{8}$$

*where $P(\cdot)$ represents the probability and $K = B_s \times c \times n$.*

*Proof*: Due to the assumption that each $e(j)$ is independent, we obtain that each $e(j)^2$ is also independent. Since $E[e(j)]=0$ and $D[e(j)]=\sigma^2$, we have the mean of each $e(j)^2$, which is expressed as

$$E[e(j)^2] = D[e(j)] + [E[e(j)]]^2 = \sigma^2, \quad j=1,...,N.$$

By invoking the *Law of Large Numbers* in probability theory, for any $\varepsilon > 0$, it leads to $\lim_{N\to\infty} P\{|\frac{1}{N}\sum_{j=1}^{N} e(j)^2 - \sigma^2| < \frac{\varepsilon}{2}\} = 1$, i.e.,

$$\lim_{N\to\infty} P\left\{\left|\frac{1}{N}\|\boldsymbol{x}-\boldsymbol{r}\|_2^2 - \sigma^2\right| < \frac{\varepsilon}{2}\right\} = 1, \tag{9}$$

Further, let $\boldsymbol{G_x}, \boldsymbol{G_r}$ denote the concatenation of all $\boldsymbol{G_{x_k}}$ and $\boldsymbol{G_{r_k}}$, $k=1,2,...,n$, respectively, and denote each element of $\boldsymbol{G_x} - \boldsymbol{G_r}$ by $\boldsymbol{G_e}(i), i=1,...,K$. Due to the assumption, we conclude that $\boldsymbol{G_e}(i)$ is independent with zero mean and variance $\sigma^2$.

Therefore, the same manipulations with Eq. (9) applied to $\boldsymbol{G_e}(i)^2$ yield $\lim_{K\to\infty} P\{|\frac{1}{K}\sum_{i=1}^{K} \boldsymbol{G_e}(i)^2 - \sigma^2| < \frac{\varepsilon}{2}\} = 1$, namely,

$$\lim_{K\to\infty} P\left\{\left|\frac{1}{K}\sum_{k=1}^{n} \|\boldsymbol{G_{x_k}} - \boldsymbol{G_{r_k}}\|_2^2 - \sigma^2\right| < \frac{\varepsilon}{2}\right\} = 1. \tag{10}$$

Considering Eqs. (10) and (9) together, we prove Eq. (8). □

According to **Theorem 1**, there exists the following equation with very large probability (limited to 1):

$$\|\boldsymbol{x}-\boldsymbol{r}\|_2^2 = \frac{N}{K}\sum_{k=1}^{n}\|\boldsymbol{G_{x_k}} - \boldsymbol{G_{r_k}}\|_2^2. \tag{11}$$

Incorporating Eq. (11) into Eq. (7) leads to

$$\begin{aligned}
&\underset{\boldsymbol{x}}{\operatorname{argmin}} \frac{1}{2}\sum_{k=1}^{n}\|\boldsymbol{G_{x_k}} - \boldsymbol{G_{r_k}}\|_2^2 + \frac{\lambda K}{N}\|\boldsymbol{\alpha_x}\|_0 \\
&= \underset{\boldsymbol{x}}{\operatorname{argmin}} \frac{1}{2}\sum_{k=1}^{n}\|\boldsymbol{G_{x_k}} - \boldsymbol{G_{r_k}}\|_2^2 + \frac{\lambda K}{N}\sum_{k=1}^{n}\|\boldsymbol{\alpha_k}\|_0 \\
&= \underset{\boldsymbol{x}}{\operatorname{argmin}} \sum_{k=1}^{n}\left(\frac{1}{2}\|\boldsymbol{G_{x_k}} - \boldsymbol{G_{r_k}}\|_2^2 + \tau\|\boldsymbol{\alpha_k}\|_0\right),
\end{aligned} \tag{12}$$

where $\tau = \lambda K/N$.

It is obvious to see that Eq. (12) can be efficiently minimized by solving $n$ sub-problems for all the groups $\boldsymbol{G_{x_k}}$, each of which is formulated below:

$$\underset{\boldsymbol{G_{x_k}}}{\operatorname{argmin}} \frac{1}{2}\|\boldsymbol{G_{x_k}} - \boldsymbol{G_{r_k}}\|_2^2 + \tau\|\boldsymbol{\alpha_k}\|_0. \tag{13}$$

Next, we will show how to learn the adaptive dictionary for each group $\boldsymbol{G_{x_k}}$, and obtain the closed-form solution for Eq. (13).

### 4.1. Adaptive Dictionary Learning

For each structural group $\boldsymbol{G_{x_k}}$, its adaptive dictionary is learned from its approximate estimate $\boldsymbol{G_{r_k}} \in \mathbb{R}^{B_S \times c}$ ($m = min(B_S, c)$).

Specifically, first, applying the singular value decomposition (SVD) to $\boldsymbol{G_{r_k}}$, that is,

$$\begin{aligned}
\boldsymbol{G_{r_k}} &= \boldsymbol{U_{r_k}} \boldsymbol{\Sigma_{r_k}} \boldsymbol{V_{r_k}}^T \\
&= \sum_{i=1}^{m} \sigma_{r_k \otimes i} \boldsymbol{u_{r_k \otimes i}} \boldsymbol{v_{r_k \otimes i}}^T,
\end{aligned} \tag{14}$$

where $\boldsymbol{\gamma_k} = [\sigma_{r_k \otimes 1}; \sigma_{r_k \otimes 2};...;\sigma_{r_k \otimes m}]$, $\boldsymbol{\Sigma_{r_k}} = \operatorname{diag}(\boldsymbol{\gamma_k})$ is a diagonal matrix with the elements of $\boldsymbol{\gamma_k}$ on its main diagonal, and $\boldsymbol{u_{r_k \otimes i}}, \boldsymbol{v_{r_k \otimes i}}$ are the columns of $\boldsymbol{U_{r_k}}$ and $\boldsymbol{V_{r_k}}$, separately.

Then, define each atom $d_{k\otimes i}$ of adaptive dictionary $D_k$ for every group $G_{x_k}$ as follows:

$$d_{k\otimes i} = u_{n_k \otimes i} v_{n_k \otimes i}^T, \ i = 1, 2, ..., m. \tag{15}$$

Finally, the learned adaptive dictionary is constructed by $D_k = [d_{k\otimes 1}, d_{k\otimes 2}, ..., d_{k\otimes m}]$.

It is clear to see that the technique of adaptive dictionary learning is very efficient, requiring only one SVD for each structural group. Experiments will demonstrate the effectiveness the designed dictionary to represent images in a sparse domain.

## 4.2. Closed-form Solution for $G_{x_k}$ Sub-problem

Before we solve each $G_{x_k}$ sub-problem (13), with the design of adaptive dictionary learning, we have the following conclusion.

**THEOREM 2.**
$$\left\| G_{x_k} - G_{n_k} \right\|_2^2 = \left\| \alpha_k - \gamma_k \right\|_2^2, \tag{16}$$

*Proof*: With the definitions about $\{\alpha_k\}$ and $\{\gamma_k\}$, we get

$$G_{x_k} = D_k \alpha_k \text{ and } G_{n_k} = D_k \gamma_k. \tag{17}$$

Then, $\left\| G_{x_k} - G_{n_k} \right\|_2^2 = \left\| D_k \alpha_k - D_k \gamma_k \right\|_2^2 = \left\| D_k (\alpha_k - \gamma_k) \right\|_2^2. \tag{18}$

Due to the construction of $D_k$ in (15) and the unitary property of $U_{n_k}$ and $V_{n_k}$, it yields

$$\begin{aligned}
\left\| D_k (\alpha_k - \gamma_k) \right\|_2^2 &= \left\| U_{n_k} \mathrm{diag}(\alpha_k - \gamma_k) V_{n_k}^T \right\|_2^2 \\
&= \mathrm{trace}\left( U_{n_k} \mathrm{diag}(\alpha_k - \gamma_k) V_{n_k}^T V_{n_k} \mathrm{diag}(\alpha_k - \gamma_k) U_{n_k}^T \right) \\
&= \mathrm{trace}\left( U_{n_k} \mathrm{diag}(\alpha_k - \gamma_k) \mathrm{diag}(\alpha_k - \gamma_k) U_{n_k}^T \right) \\
&= \mathrm{trace}\left( \mathrm{diag}(\alpha_k - \gamma_k) U_{n_k}^T U_{n_k} \mathrm{diag}(\alpha_k - \gamma_k) \right) \\
&= \mathrm{trace}\left( \mathrm{diag}(\alpha_k - \gamma_k) \mathrm{diag}(\alpha_k - \gamma_k) \right) = \left\| \alpha_k - \gamma_k \right\|_2^2.
\end{aligned} \tag{19}$$

Combining Eqs. (18) and (19), we prove Eq. (16). □

With the aid of **Theorem 2**, the sub-problem (13) is equivalent to

$$\underset{\alpha_k}{\mathbf{argmin}} \ \tfrac{1}{2} \left\| \alpha_k - \gamma_k \right\|_2^2 + \tau \left\| \alpha_k \right\|_0. \tag{20}$$

Owing to Lemma 2 in [10], the closed-form solution of (20) is expressed as

$$\hat{\alpha}_k = \mathrm{hard}(\gamma_k, \sqrt{2\tau}) = \gamma_k \cdot \mathbf{1}(\mathrm{abs}(\gamma_k) - \sqrt{2\tau}), \tag{21}$$

where $\mathrm{hard}(\cdot)$ denotes the operator of hard thresholding and $\cdot$ stands for the element-wise product of two vectors. Thus, the efficient solution for sub-problem (13) is

$$\hat{G}_{x_k} = D_k \hat{\alpha}_k. \tag{22}$$

This process is applied for all groups to achieve estimates $\hat{G}_{x_k}$ and all $\hat{G}_{x_k}$ are returned to their original positions and averaged at each pixel to obtain the solution for Eq. (7).

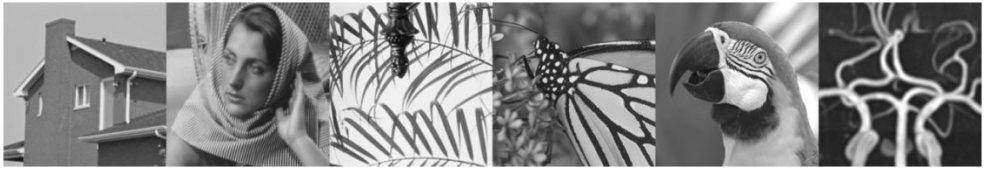

**Figure 1:** Experimental test images.

## 4.3. Summary

In light of all derivations above, the complete description of image CS recovery via structural group sparse representation (SGSR) is given below:

---
**Image CS Recovery via SGSR**

---
**Input:** The observed measurement $b$, the measurement matrix $A$ and parameter $\lambda$;
  Initialization: set initial estimate $x^{(0)}$;
  **for** Iteration number $j = 0, 1, 2, \ldots$, Max_iter
    Update $r$ by computing (6);
    Create structural groups $G_{r_k}$ by searching similar patches in new $r$;
    **for** Each group $G_{r_k}$
      Construct dictionary $D_k$ by computing Eq. (15);
      Reconstruct $\hat{G}_{x_k}$ by computing Eq. (22);
    **end for**
    Update $x$ by weighted averaging all the reconstructed groups $\hat{G}_{x_k}$;
  **end for**
**Output:** Final recovered image $\hat{x}$.

---

## 5. Experimental Results

In this section, experimental results are presented to evaluate the performance of the proposed CS recovery via SGSR. Six test images are shown in Figure 1. In our experiments, the CS measurements are obtained by applying a Gaussian random projection matrix to the original image signal at block level, i.e., block-based CS with block size of 32×32 [16]. SGSR is compared with five CS recovery methods in literature, i.e., wavelet method (DWT), contourlet method (CT), total variation (TV) method[1] [8], multi-hypothesis (MH) method[2] [18], collaborative sparsity (CoS) method[3] [21], which deal with image signals in the wavelet domain, contourlet domain [19], gradient domain, random projection residual domain, and hybrid space-transform domain, respectively. It is worth emphasizing that MH and CoS are known as the state-of-the-art algorithms for image CS recovery.

**Table 1:** PSNR comparisons with different CS recovery methods (dB)

| Ratio | Algorithms | House | Barbara | Leaves | Monarch | Parrots | Vessels | Avg. |
|---|---|---|---|---|---|---|---|---|
| 20% | DWT | 30.70 | 23.96 | 22.05 | 24.69 | 25.64 | 21.14 | 24.70 |
| | CT | 30.06 | 24.20 | 21.45 | 24.83 | 25.97 | 21.48 | 24.66 |
| | TV | 31.44 | 23.79 | 22.66 | 26.96 | 26.68 | 22.04 | 25.59 |
| | MH | 33.60 | 31.09 | 24.54 | 27.03 | 28.06 | 24.95 | 28.21 |
| | CoS | 34.34 | 26.60 | 27.38 | 28.65 | 28.44 | 26.71 | 28.69 |
| | **SGSR** | **35.70** | **33.45** | **28.61** | **29.01** | **29.84** | **29.83** | **31.07** |
| 30% | DWT | 33.60 | 26.26 | 24.47 | 27.23 | 28.03 | 24.82 | 27.40 |
| | CT | 32.32 | 25.58 | 23.75 | 27.24 | 27.88 | 24.18 | 26.82 |
| | TV | 33.75 | 25.03 | 25.85 | 30.01 | 28.71 | 25.13 | 28.08 |
| | MH | 35.54 | 33.47 | 27.65 | 29.18 | 31.20 | 29.36 | 31.07 |
| | CoS | 36.69 | 29.49 | 31.02 | 31.38 | 30.39 | 31.35 | 31.72 |
| | **SGSR** | **37.34** | **35.83** | **32.75** | **32.09** | **33.15** | **34.06** | **34.20** |
| 40% | DWT | 35.69 | 28.53 | 26.82 | 29.58 | 30.06 | 29.53 | 30.03 |
| | CT | 34.33 | 27.25 | 25.74 | 29.51 | 29.85 | 27.10 | 28.96 |
| | TV | 35.56 | 26.56 | 28.79 | 32.92 | 30.54 | 28.14 | 30.42 |
| | MH | 37.04 | 35.20 | 29.93 | 31.07 | 33.21 | 33.49 | 33.32 |
| | CoS | 38.46 | 32.76 | 33.87 | 33.98 | 32.55 | 33.95 | 34.26 |
| | **SGSR** | **39.09** | **37.74** | **35.78** | **34.80** | **35.59** | **37.57** | **36.76** |

---
[1] http://www.caam.rice.edu/~optimization/L1/TVAL3/.
[2] http://www.ece.msstate.edu/~fowler/BCSSPL/.
[3] http://idm.pku.edu.cn/staff/zhangjian/RCoS/.

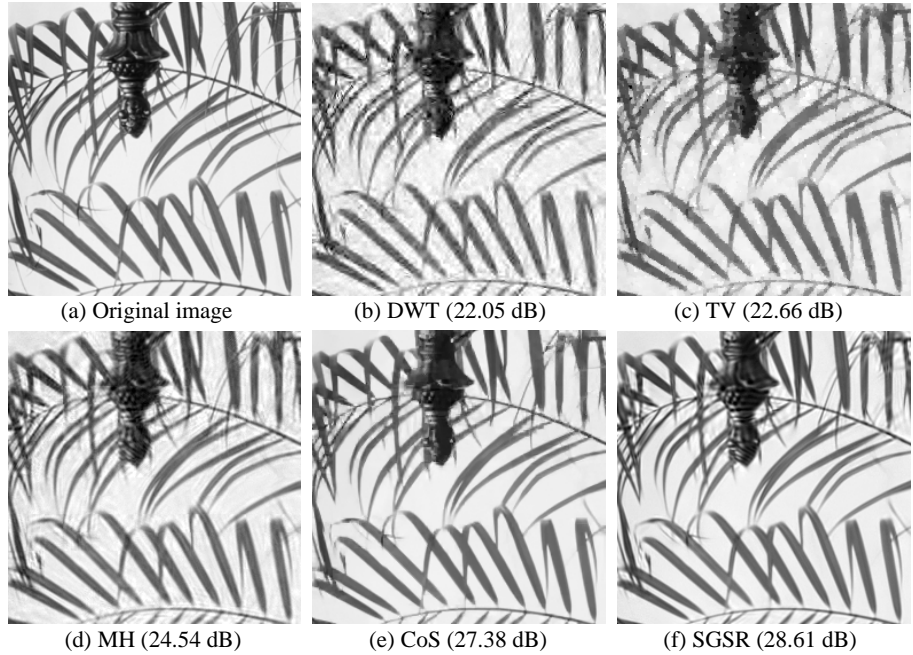

(a) Original image     (b) DWT (22.05 dB)     (c) TV (22.66 dB)

(d) MH (24.54 dB)     (e) CoS (27.38 dB)     (f) SGSR (28.61 dB)

**Figure 2:** Visual comparison of CS recovered results for *Leaves* by different methods (ratio = 20%).

In our implementation, all the parameters of SGSR are set empirically for all test images. Concretely, the size of each patch, i.e., $\sqrt{B_s} \times \sqrt{B_s}$ is set to be 8×8, the size of training window for searching matched patches, i.e., $L \times L$ is set to be 40×40, and the number of best matched patches, i.e., $c = 50$ for each group, and $\lambda = 1.8e3$, $\rho = 1$. It is necessary to stress that the choice for all the parameters is general, and can be generalized to other natural images, which has been verified in our experiments. In this paper, we exploit the results of MH as initialization of the proposed SGSR for image CS recovery.

The PSNR comparisons for all the test images in the cases of 20% to 40% measurements are provided in Table 1. SGSR provides quite promising results, achieving the highest PSNR among the six comparative algorithms over all the cases, which can improve roughly 7.2 dB, 6.6 dB, 6.0 dB, 3.1 dB and 2.4 dB on average, compared with CT, DWT, TV, MH, and CoS, respectively.

Some visual results of the recovered images by various algorithms are presented in Figures 2–3. Obviously, DWT and TV generate the worst perceptual results. The CS recovered images by MH and CoS possess much better visual quality than those of DWT and TV, but still suffer from some undesirable artifacts, such as ringing effects and lost details. The proposed algorithm SGSR not only eliminates the ringing effects, but also preserves sharper edges and finer details, showing much clearer and better visual results than the other competing methods. The high performance of SGSR is attributed to the proposed adaptive group sparse representation modeling, which offers a powerful mechanism of characterizing the structured sparsities of natural image signals.

Because the objective function (5) is non-convex, it is difficult to give its theoretical proof for global convergence. Here, we only provide empirical evidence to illustrate the nice convergence of the proposed CS recovery scheme. Figure 4 plots the evolutions of PSNR versus iteration numbers for four test images with various ratios of measurements. It is observed that with the growth of iteration number, all the PSNR curves increase monotonically and ultimately become flat and stable.

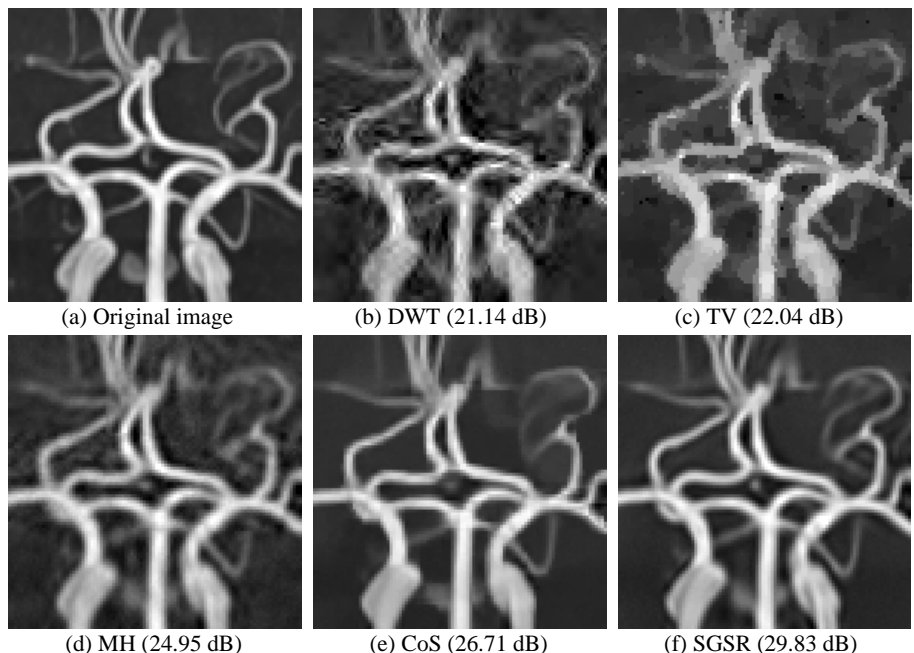
(a) Original image    (b) DWT (21.14 dB)    (c) TV (22.04 dB)
(d) MH (24.95 dB)    (e) CoS (26.71 dB)    (f) SGSR (29.83 dB)

**Figure 3:** Visual comparison of CS recovered results for *Vessels* by different methods (ratio = 20%).

The complexity of SGSR is provided as follows. Assume that the number of image pixels is $N$, that the average time to compute similar patches for each reference patch is $T_s$. The SVD of each group $G_{n_k}$ with size of $B_s \times c$ is $\mathcal{O}(B_s \times c^2)$. Hence, the total complexity of SGSR is $\mathcal{O}(N(B_s c^2 + T_s))$. For a 256×256 image, the proposed algorithm SGSR and CoS [21] requires about 6~7 minutes for CS recovery on an Intel Core2 Duo 2.96G PC under Matlab R2011a environment, while the other four comparative methods require about 1~3 minutes.

## 6. Conclusion

In this paper, a novel sparse representation modeling, called structural group sparse representation (SGSR), is proposed, which efficiently characterizes the intrinsic sparsity and self-similarity of natural images in an adaptive group domain. Extensive experiments manifest that the developed CS recovery strategy via SGSR is able to increase image recovery quality by a large margin compared with the current existing methods. Ongoing work includes the extensions on a variety of other image restoration applications by taking advantage of the proposed SGSR.

## Acknowledgement

The authors would like to thank the anonymous reviewers for their helpful and constructive suggestions. This work was supported in part by National Basic Research Program of China (973 Program 2009CB320905) and by the National Science Foundation of China under Grants 61272386 and 61100096.

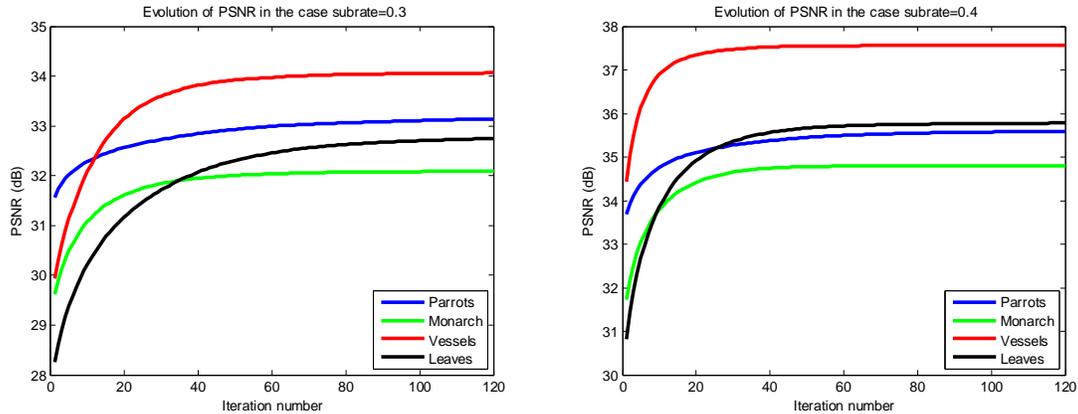

**Figure 4:** Progression of the PSNR (dB) results achieved by proposed SGSR for four test images with respect to the iteration number in the cases of ratio=30% and ratio=40%.